\documentclass[10pt,twocolumn,letterpaper]{article}

\usepackage[pagenumbers]{cvpr} % To force page numbers, e.g. for an arXiv version
\usepackage{multirow} 
\usepackage{tabularx}
\usepackage{booktabs}
\usepackage{subcaption}
\usepackage{xurl}
\usepackage[accsupp]{axessibility} 
\usepackage{makecell}
\definecolor{cvprblue}{rgb}{0.21,0.49,0.74}
\usepackage[pagebackref,breaklinks,colorlinks,allcolors=cvprblue]{hyperref}

\def\paperID{26} % *** Enter the Paper ID here
\def\confName{CVPR}
\def\confYear{2026}

\title{Scene-Agnostic Object-Centric Representation Learning for 3D~Gaussian~Splatting}

\author{Tsuheng Hsu$^1$ \quad Guiyu Liu$^2$ \quad Juho Kannala$^{1,2}$ \quad Janne Heikkilä$^2$\\
$^1$Aalto University \quad $^2$University of Oulu\\
}

\begin{document}
\maketitle
\begin{abstract}
Recent works on 3D scene understanding leverage 2D masks from visual foundation models (VFMs) to supervise radiance fields, enabling instance-level 3D segmentation. However, the supervision signals from foundation models are not fundamentally object-centric and often require additional mask pre/post-processing or specialized training and loss design to resolve mask identity conflicts across views. The learned identity of the 3D scene is scene-dependent, limiting generalizability across scenes. Therefore, we propose a dataset-level, object-centric supervision scheme to learn object representations in 3D Gaussian Splatting (3DGS). Building on a pre-trained slot attention-based Global Object Centric Learning (GOCL) module, we learn a scene-agnostic object codebook that provides consistent, identity-anchored representations across views and scenes. By coupling the codebook with the module's unsupervised object masks, we can directly supervise the identity features of 3D Gaussians without additional mask pre-/post-processing or explicit multi-view alignment. The learned scene-agnostic codebook enables object supervision and identification without per-scene fine-tuning or retraining. Our method thus introduces unsupervised object-centric learning (OCL) into 3DGS, yielding more structured representations and better generalization for downstream tasks such as robotic interaction, scene understanding, and cross-scene generalization.
\end{abstract}    
\section{Introduction}
\label{sec:intro}

\label{sec:formatting}
\begin{figure*}[t]
  \centering
  \includegraphics[width=1\linewidth]{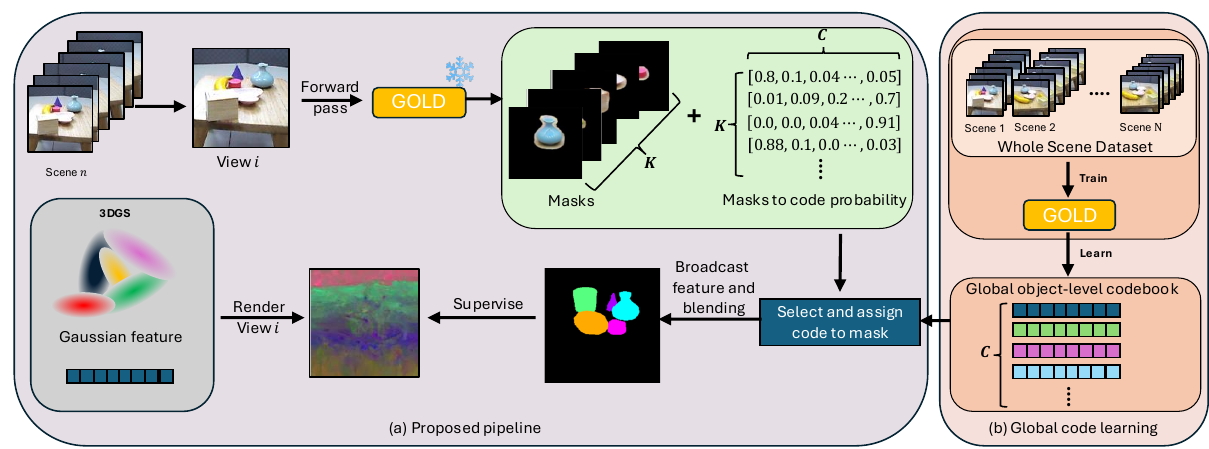}
  \caption{Overview of our proposed pipeline. (b) GOLD learns a global, scene-agnostic codebook storing a total of $C$ object intrinsic features while preserving the underlying functionality of slot attention.During scene optimization (a), for each view, we render Gaussian identity features and feed the input image into the trained GOLD module to obtain $K$ object masks and mask-to-codebook probabilities. We assign a codebook feature to each mask, broadcast it to image resolution, and blend the masks to construct dense supervision for Gaussian identity features.}
  \label{fig:pipeline}
\end{figure*}

Object-level 3D scene understanding is crucial for downstream applications such as embodied AI, robotics, and scene editing, which require scenes to be understood in terms of interactable objects. Recent advances in 3D representations, including radiance field methods such as Neural Radiance Fields (NeRF) \cite{barron2021mip,mildenhall2021nerf} and 3D Gaussian Splatting (3DGS) \cite{kerbl20233d,lu2024scaffold}, provide high-fidelity and efficient 3D representations that enable high-quality reconstructions and real-time rendering; however, their lack of semantic understanding hinders downstream applications.

To address this issue, various works leverage the segmentation capability of 2D Vision Foundation Models (VFM) \cite{ren2024grounded,kirillov2023segment,carion2025sam} by lifting their masks into 3D to supervise 3D representations \cite{cen2025segmentnerf,kim2024garfield,ye2024gaussian,cen2025segment,piekenbrinck2025opensplat3d,zhu2025rethinking,zhu2025objectgs}, enabling 3D segmentation and open-vocabulary understanding. However, VFM-generated masks lack consistency across views; e.g., the same objects across views have different segmentation IDs at different angles of view, thereby counteracting the supervision. Therefore, additional pre-/post-processing of the mask or pipeline is required; some methods \cite{ye2024gaussian,zhu2025objectgs} leverage tracking foundation models such as DEVA \cite{cheng2023tracking} for tracking and aligning masks across views. Beyond tracking-based solutions, other approaches rely on contrastive learning, introducing additional loss terms and tailored training strategies to enforce multi-view consistency \cite{zhu2025rethinking, cen2025segment, ying2024omniseg3d}, allowing direct supervision but increasing training instability and the amount of data needed.

Another limitation of VFMs is that they are trained to perform large-scale, appearance-driven segmentation of arbitrary image regions, rather than reasoning about scenes as compositions of consistent and persistent objects. As a result, their predictions primarily reflect local visual similarity, often segmenting regions based on surface-level cues without capturing the underlying object structure or relationships between different parts of an object, leading to over-segmentation of single physical entities, which will significantly undermine applications that require physical interaction of the entities in the scene \cite{li2024object, chen2026semantic}. These observations motivate the need for object-centric representations that are discovered in a scene-agnostic way, rather than relying on per-scene, appearance-based VFM masks.

In contrast to VFM-based mask supervision, a growing line of work explores unsupervised object-centric learning (OCL), where objects are discovered directly from data without manual labels. Slot attention-based works \cite{locatello2020object, jia2022improving, seitzer2022bridging} facilitate unsupervised object-centric discovery, and are applied to NeRF, achieving unsupervised 3D OCL \cite{yu2021unsupervised, luo2024unsupervised, smith2022unsupervised, stelzner2021decomposing}. However, due to the implicit representation of NeRF, it is non-trivial for further downstream applications. While 3DGS emerges as a compelling alternative to NeRF due to its explicitness and efficiency, the slot attention-based method is non-trivial. In particular, slot permutation leads to inconsistent object representations across views, making it difficult to provide stable object-centric supervision.

To address these challenges, we deliberately build on recent work in global object-centric learning (GOCL), and, in particular, adopt Global Object-centric Learning via Disentangled Slot Attention (GOLD) \cite{chen2025learning} as the backbone. With the learned scene-agnostic codebook and the unsupervised object masks from GOLD, we show that unsupervised OCL can be introduced into 3DGS, yielding object-centric 3D representations whose identities are anchored in a global codebook, thereby enabling cross-scene identification within compact pipeline without the need for mask pre-processing or tailored training and loss design.

Our main contributions are:
\begin{itemize}
    \item We integrate unsupervised object-centric learning with 3D Gaussian Splatting for the first time.
    \item A scene-agnostic object codebook to provide consistent object identities across views and scenes.
    \item A supervision pipeline that avoids VFM-specific pre-/post-processing of masks and does not rely on additional view-consistency losses or ad hoc training strategies.
\end{itemize}
Our implementation is available at \url{https://github.com/HenryHsu0217/SA-OCL}.
\iffalse
\begin{table*}[!t]
\centering
\caption{Comparison of our pipeline with prior VFM-based methods.}
\label{tab:comparison}
\small
\renewcommand{\arraystretch}{1.15} 
% Using tabular* to stretch across the full page width evenly
\begin{tabular*}{\textwidth}{@{\extracolsep{\fill}} l c c c c c c @{}}
\toprule
\textbf{Feature Learning Method} & \textbf{Pre/post-proc.} & \textbf{Training} & \textbf{Views} & \textbf{Supervision} & \textbf{Obj-centric} & \textbf{Identity} \\
\midrule
Post-hoc~\cite{ye2024gaussian, li2024object} & Required & Standard & Standard & 2D VFM masks & Not guaranteed & Scene-dependent \\
End-to-end ~\cite{zhu2025rethinking, cen2025segment} & None & Complex & Dense & 2D VFM masks & Not guaranteed & Scene-dependent \\
\midrule
\textbf{Ours} & \textbf{None} & \textbf{Standard} & \textbf{Standard} & \textbf{\makecell{Global codebook + \\ object-centric masks}} & \textbf{Yes} & \textbf{Cross-scene} \\
\bottomrule
\end{tabular*}
\end{table*}
\fi

\section{Related Work}
\subsection{Radiance field 3D Scene Representation}
Radiance field representation of 3D scenes has emerged as a superior alternative to classic representations due to its ability to be constructed from 2D images and to model complex geometry and view-dependent appearance in a continuous, photorealistic manner. NeRF \cite{mildenhall2021nerf} initiated a surge of work on neural scene representations by modeling the radiance field with neural networks. However, due to its implicit representation, NeRF is limited in supporting certain downstream applications. In contrast, 3D Gaussian Splatting (3DGS)  \cite{kerbl20233d,cheng2024gaussianpro, lu2024scaffold} represents the radiance field using explicit 3D Gaussians, enabling a wider range of applications, including 4D scene scene modeling \cite{wu20244d,ren2023dreamgaussian4d}, autonomous driving \cite{zhou2024drivinggaussian}, and robotics \cite{lu2024manigaussian,li2024object}. Therefore, we adopt 3DGS as the primary representation in our pipeline.

\subsection{3D Semantic Learning with Neural Scene Representations}
The lack of semantic and instance-level information in 3D neural scene representation has driven the development of learning-based methods that incorporate semantic supervision into radiance field models. However, high-quality 3D annotated data is limited; therefore, VFMs such as SAM \cite{kirillov2023segment} and its subsequent works \cite{ravi2024sam,carion2025sam,ren2024grounded}, which generate high-quality 2D segmentation mask are widely adopted as supervision. Recent work \cite{piekenbrinck2025opensplat3d, ye2024gaussian, zhu2025rethinking, cen2025segment, zhu2025objectgs, choi2024click, qin2024langsplat} distill 2D VFM features from SAM mask or DINO \cite{caron2021emerging} directly to 3D Gaussians for 3D scene understanding, grouping, and segmentation. 

For example, Gaussian Grouping \cite{ye2024gaussian} proposed a pipeline to supervise the identity feature of Gaussians with SAM masks utilizing DEVA \cite{cheng2023tracking} to resolve the mask contrastive issue and directly supervise the Gaussian identity feature. ObjectGS \cite{zhu2025objectgs} focuses on object-level 3DGS segmentation optimization by projecting and voting from a pre-processed SAM mask, assigning one-hot object IDs to the point clouds, and employs Scaffold-GS-based \cite{lu2024scaffold} densification control to maintain per-instance consistency. In contrast to prior paradigms, Unified Lift \cite{zhu2025rethinking} proposed a pipeline that is pre- and post-processing free by treating the SAM mask as a noisy label and applying contrastive loss for coarse identity feature learning and then learn a scene-dependent codebook through association mapping, concentration loss, and label filtering for identification. Compared to our global codebook, their codebook lacks cross-scene generalization for either supervising or identification.

However, relying on VFM masks not only requires additional design or processing to address mask-label inconsistencies across views but also leads to limited object-centricity and poor cross-scene identity generalization. Therefore, unlike prior work, our proposed pipeline delivers both object-centric results and cross-scene generalization without any pre-/post-processing of the supervision source, nor any re-training, fine-tuning, or explicit optimization of identity properties.

\subsection{Unsupervised Object-Centric Learning in 3D}
Object-Centric Learning (OCL) aims to enable models to perceive, reason, and generalize about the world in terms of composable entities without supervision. Slot Attention models \cite{locatello2020object,elsayed2022savi++, jia2022improving, zadaianchuk2023object} discover objects by using attention to group visual features into slots and learn slot latents that encode entities. Motivated by the compositional inductive bias of Slot Attention, prior works \cite{yu2021unsupervised, luo2024unsupervised, smith2022unsupervised, stelzner2021decomposing} extend slot-based architectures to 3D scenes with NeRF representations, yielding object-centric neural fields. Specifically, \cite{stelzner2021decomposing, smith2022unsupervised, yu2021unsupervised} apply Slot Attention to input views and condition NeRF decoders on the learned slot latents and novel-view parameters to obtain 3D object-centric scene representations. In contrast, \cite{luo2024unsupervised} introduces translation invariance by disentangling object location from object latent codes, projecting 2D object locations into 3D space, and decoding in object-local coordinates.

Despite the success of unsupervised 3D OCL in NeRF, downstream 3D applications remain limited by NeRF’s implicit representation. At the same time, supervising 3DGS with slot attention-based OCL is non-trivial due to slot permutation. Recent work extends slot-based models to video settings to mitigate slot permutation by enforcing temporal consistency \cite{manasyan2025temporally} and leveraging motion cues \cite{kipf2021conditional, elsayed2022savi++,zadaianchuk2023object}. Video-based Slot Attention stabilizes slot identities over time, making it possible to associate object representations consistently across views. However, although temporal consistency alleviates slot permutation, these approaches—similar to VFMs—typically operate within single scenes and entangle object appearance with scene-specific factors, limiting cross-scene generalization.

To improve generalization, recent GOCL methods, such as GOLD \cite{chen2025learning} and CSGO \cite{chen2025unsupervised}, disentangle intrinsic object properties from extrinsic factors such as pose and lighting. GOLD formulates this within a variational end-to-end framework, whereas CSGO adopts a multi-stage training paradigm, enabling object-centric representations that remain consistent across scenes.

Building on these advances, we extend OCL to explicit 3D scene representations. Specifically, we integrate GOCL into 3DGS, enabling cross-scene object identification and generalization within an explicit radiance field representation.

\section{Method}

The overall proposed pipeline is shown in Fig.~\ref{fig:pipeline}. In Sec.~\ref{sec:gold_training}, we first describe the training of our GOLD \cite{chen2025learning} backbone, where the global object-level codebook is learned for supervision. Sce.~\ref{sec:gau_render} describes how our 3D Gaussian is rendered and the identity feature is learned. In Sec.~\ref{sec:supervise}, we show how the learned object-level codebook provides global, scene-agnostic object supervision for object-centric 3D feature learning and identification. 

\subsection{Global Codebook Learning}
\label{sec:gold_training}
We adopt the GOLD \cite{chen2025learning} framework to learn global, scene-agnostic object representations. Similarly to standard slot attention models, we utilize DINO \cite{caron2021emerging} features as both input and reconstruction target. The representation learning process operates as follows:
\begin{itemize}
    \item Background Encoding: A background encoder processes the extracted DINO  features into a latent background slot, $s^{bck} \in \mathbb{R}^{D_{slot}}$, where $D_{slot}$ is the overall slot dimension.
    \item Global Codebook Setup: In parallel, a Disentangled Slot Attention (DSA) module maintains an independently initialized global codebook, $e^{glo} \in \mathbb{R}^{C \times D_{code}}$, where $C$ is the total global object count and $D_{code}$ is the dimension of the codebook features.
    \item Slot Initialization: For $K$ object slots inside the DSA, instead of re-initializing the entire slot representation per forward pass, only the scene-dependent extrinsic features $s^{ext} \in \mathbb{R}^{K \times D_{ext}}$ and identity logits $\gamma \in \mathbb{R}^{K \times C}$ is initialized from learnable Gaussian distributions.
    \item Intrinsic Representation: To force the model to select a discrete global prototype for each object during attention in the DSA, we compute an intrinsic representation $s^{int} \in \mathbb{R}^{K \times D_{code}}$ via a Gumbel-Softmax \cite{jang2016categorical} over the logits:
    \begin{equation}
        s^{int} = \text{Gumbel-Softmax}(\gamma) \cdot e^{glo}.
    \end{equation}
    
    \item Full Slot Construction \& Refinement: The intrinsic, extrinsic, and background components are concatenated to construct the final representations for iterative attention inside DSA:
     \begin{equation}
         s^{full} = [ [s^{int}, s^{ext}] ; s^{bck} ] \in \mathbb{R}^{(K+1) \times D_{slot}}.
     \end{equation} 
     During this iterative attention process, $s^{ext}$, $s^{int}$, and $\gamma$ are refined jointly.
\end{itemize}
This allows each slot to align with specific scene features while consistently selecting and encoding global object characteristics. The model is then trained by decoding and blending the refined slots, using a DINO feature reconstruction loss together with variational regularization terms. This design effectively disentangles scene-dependent attributes (e.g., pose and scale) from scene-invariant object identity, yielding a compact and transferable representation for cross-scene understanding and supervision.
\subsection{Learning Gaussian Identity features}
\label{sec:gau_render}
With the learned global codebook and GOLD’s unsupervised object masks, we eliminate the need for mask pre-/post-processing and additional Gaussian-level feature learning modules. We therefore follow the compact pipeline of Gaussian Grouping \cite{ye2024gaussian} on top of the 3DGS backbone \cite{kerbl20233d}. The goal is to learn a set of 3D Gaussians $G_i=\{g_i\}^N_{i=1}$, where $g_i= \{p_i, s_i, q_i, o_i, c_i \}$ contains trainable parameters, which $p_i$ denotes the center of the Gaussian, $s_i, q_i$ denotes the scale and quaternion respectively, $o_i$ and $c_i$ is the opacity value and color values, which the color value is modeled by Spherical Harmonic (SH) coefficients. With the proposed tile-based rasterization \cite{kerbl20233d}, each Gaussian function $G_i(x)$ is projected to the 2D image plane, resulting in 2D Gaussians $G'(u)$. Then, for each pixel $u$ in the 2D image plane, the associated Gaussians are sorted by distance, and $\alpha$-blending is applied to compute the color $C_u=\sum_{i\in \mathcal{N}}c_i\alpha_i\prod_{t=1}^{i-1}(1-\alpha_i)$, where $a_i$ is the occupancy of the projected Gaussian: $o_i\cdot G'_i(u)$ and $ \mathcal{N}$ is the total Gaussian that is associated to pixel $u$.

To learn Gaussian-level identity features, an identity feature $f_i$ of size $D_{code}$ is attached to each Gaussian. Following \cite{ye2024gaussian}, the identification features are rendered onto the 2D image plane, similar to the  color rendering process, with the SH degree set to 0 as:
\begin{equation}
    F_{u} = \sum_{i\in  \mathcal{N}}f_i\alpha_i\prod_{t=1}^{i-1}(1-\alpha_i).
\end{equation}
With each pixel attached to an identity feature, we then further supervise it with our learned global codebook.
\subsection{Global Object-Level Supervision}
\label{sec:supervise}
We supervise the Gaussian identity feature during per-view optimization of a scene directly. During optimization of a scene, giving input view image $I_i, i\in V$ where $V$ is the total number of supervising views of a scene, $I_i$ is passed into the pre-trained GOLD module, outputting $K$ image masks $\mathcal{M}=\{ M_k\}_{k=1}^K$, and probability matrix $\gamma\in\mathbb{R}^{K\times C}$. For each mask $k\in\{1\dots K\}$, we first identify the index $c^*_k$ associated with the maximum probability and assign the corresponding feature vector from the global codebook to the $k$-th mask:
\begin{equation}
    c^*_k=\arg\max_{j\in\{1\dots, C\}} \gamma_{k,j}, \quad f_k^*=e^{glo}_{c^*_k}\in \mathbb{R}^{D_{code}}.
\end{equation}
For each selected codebook feature ${f}_k^*$, we broadcast it to match the spatial resolution of the mask $M_k$ and aggregate them into a dense target feature map:
\begin{equation}
F^*_u = \sum_{k=1}^{K} M_k(u) \cdot f_k^*.
\end{equation}

Without requiring an additional linear layer (as in \cite{ye2024gaussian}), we directly supervise the rendered Gaussian identity features using an MSE loss:

\begin{equation}
    \mathcal{L}_{\text{feature}} = \frac{1}{|\Omega|} \sum_{u \in \Omega} \left\| F_u - F_u^* \right\|_2^2
\end{equation}
where $\Omega$ is the set of all pixel samples. Similar to \cite{ye2024gaussian}, we include 3D Regularization Loss for better grouping of Gaussians, which randomly samples $m$ Gaussians in the scene and computes the KL divergence between the class probability distribution of each sampled Gaussian and those of its top-$k$ nearest neighbors. For a given Gaussian $G_i$, its class probability $p_{i,c}, \ c\in C$ is calculated by cosine similarity between the codebook features, followed by softmax as:
\begin{equation}
s_c = \frac{f_i \cdot e^{glo}_c}{|f_i|_2 |e^{glo}_c|_2},\quad p_{i,c} = \frac{\exp(s_c)}{\sum_{j=1}^{C} \exp(s_j)}.
\end{equation}
We then formalize our 3D Regularization Loss as:
\begin{equation}
    \mathcal{L}_{3D}=\frac{1}{mk}\sum_{i=1}^m\sum_{j=1}^k\sum_{c=1}^Cp_{i,c}\log\left(\frac{p_{i,c}}{p_{j,c}}\right)
\end{equation}
where $k$ denotes the number of nearest neighbors.

With the feature loss $\mathcal{L}_{feature}$ and 3D Regularization loss $\mathcal{L}_{3D}$, combining with the image rendering loss \cite{kerbl20233d} $\mathcal{L}_{rend}$, the complete training loss is
\begin{equation}
    \mathcal{L}=\mathcal{L}_{rend} + \lambda_{feature}\mathcal{L}_{feature}+ \lambda_{3D}\mathcal{L}_{3D}.
\end{equation}
During rendering and inferencing, we simply assign the code index to each pixel by calculating the $L2$ distance between the rendered feature and the codebook feature as:
\begin{equation}
    ID_u=\arg\min_{c\in \{1\dots C\}}  \left\| F_u - e_c^{glo} \right\|_2^2.
\end{equation}
\section{Experiment}

\begin{figure*}[t]
  \centering
  \includegraphics[width=0.95\textwidth]{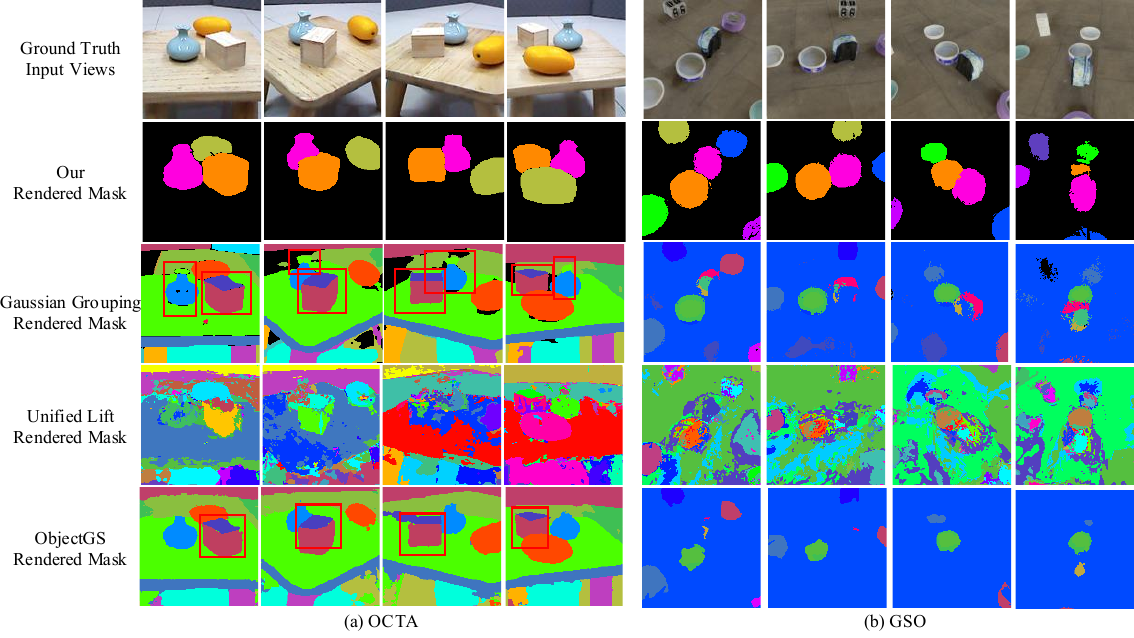}
  \caption{Qualitative comparison of rendered Gaussian feature masks on the OCTA dataset (left) and GSO dataset (right), where each object ID is visualized with a distinct color generated through a deterministic mapping. VFM-based supervision lacks object-level consistency, as highlighted in the images, where different surfaces of the cube or the rim of the vessel are often identified as separate instances.}
  \label{fig:OverallVis}
\end{figure*}

\begin{table*}[t]
\centering
\caption{Quantitative comparison of our method and prior works on the OCTA and GSO datasets, evaluated across test scenes. Best results are in \textbf{bold}, second best are underlined.}
\label{tab:mainmetric}

\resizebox{0.8\textwidth}{!}{
\begin{tabular}{llccccc}
\toprule
Dataset & Method & FG-ARI $\uparrow$ & ARI-A $\uparrow$ & FG-AMI $\uparrow$ & AMI-A $\uparrow$ & mIoU $\uparrow$ \\
\midrule

\multirow{4}{*}{OCTA}

& Gaussian Grouping \cite{ye2024gaussian} & \underline{0.700 ± 0.146} & \underline{0.149 ± 0.042} & \underline{0.751 ± 0.142} & \underline{0.411 ± 0.076} & \textbf{0.697 ± 0.074} \\
& Unified Lift \cite{zhu2025rethinking} & 0.474 ± 0.106 & 0.104 ± 0.027 & 0.567 ± 0.113 & 0.309 ± 0.060 & 0.500 ± 0.062 \\
& ObjectGS \cite{zhu2025objectgs} & 0.586 ± 0.158 & 0.126 ± 0.057 & 0.642 ± 0.141 & 0.343 ± 0.087 & 0.466 ± 0.139 \\
& Ours & \textbf{0.870 ± 0.061} & \textbf{0.760 ± 0.032} & \textbf{0.868 ± 0.042} & \textbf{0.677 ± 0.030} & \underline{0.593 ± 0.079} \\
\midrule

\multirow{4}{*}{GSO}

& Gaussian Grouping \cite{ye2024gaussian} & \underline{0.549 ± 0.211} & \underline{0.554 ± 0.275} & \underline{0.651 ± 0.232} & \underline{0.557 ± 0.180} & \textbf{0.594 ± 0.084} \\
& Unified Lift \cite{zhu2025rethinking} & 0.292 ± 0.110 & 0.197 ± 0.103 & 0.478 ± 0.174 & 0.278 ± 0.083 & 0.396 ± 0.049 \\
& ObjectGS \cite{zhu2025objectgs} & 0.396 ± 0.225 & 0.342 ± 0.221 & 0.529 ± 0.207 & 0.341 ± 0.153 & 0.314 ± 0.118 \\
& Ours & \textbf{0.846 ± 0.208} & \textbf{0.644 ± 0.147} & \textbf{0.845 ± 0.204} & \textbf{0.558 ± 0.130} & \underline{0.508 ± 0.130} \\
\bottomrule
\end{tabular}
}
\end{table*}

\begin{figure*}[t]
  \centering
  \begin{subfigure}[t]{0.49\textwidth}
    \centering
    \includegraphics[width=\linewidth]{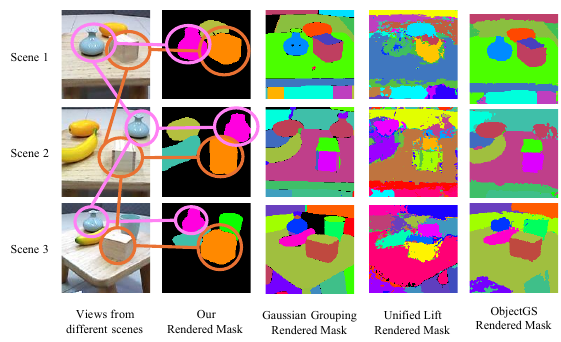}
    \caption{OCTA Dataset}
    \label{fig:top}
  \end{subfigure}
 \hfill
  \begin{subfigure}[t]{0.49\textwidth}
    \centering
    \includegraphics[width=\linewidth]{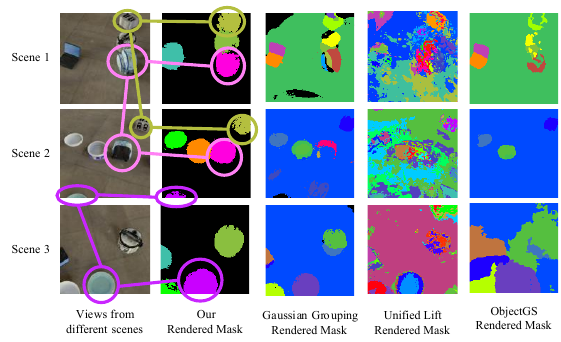}
    \caption{GSO Dataset}
    \label{fig:bottom}
  \end{subfigure}

  \caption{Visualization of cross-scene identity on the OCTA dataset (a) and GSO dataset (b). Our method achieves cross-scene identification using the learned global codebook: as highlighted, the same object is assigned a consistent ID across different scenes, whereas VFM supervised methods remain scene-dependent. }
  \label{fig:cross_scene}
\end{figure*}

\subsection{Dataset}
We evaluate our pipeline on two datasets that are compatible with both 3DGS and OCL tasks, OCTScene-A (OCTA) \cite{huang2023octscenes} and Google Scanned Objects (GSO) \cite{downs2022google}. OCTA contains 3100 train scenes and 100 test scenes of real-world tabletop scenes captured from 60 viewpoints over a $360^\circ$ viewing range. The GSO dataset is generated using Kubric \cite{greff2022kubric}, where 1900 training scenes and 90 test scenes are rendered using the same camera setup as OCTA. Both datasets feature 1 to 7 different objects per-scene and 11 total different objects. 

\subsection{Experiment Setup}
\subparagraph{Comparison}
We compare our pipeline with representative approaches for 3D scene feature learning under VFM-based 3DGS frameworks. We consider Gaussian Grouping \cite{ye2024gaussian} as a foundational approach for Gaussian identity feature learning with explicit 3D Gaussians, which relies on 2D mask projection with pre-processing. We further compare with ObjectGS \cite{zhu2025objectgs}, which extends this paradigm to object-level segmentation but still depends on pre-processed mask supervision. In contrast, Unified Lift \cite{zhu2025rethinking} proposes an end-to-end training pipeline without pre-/post-processing, learning scene-dependent object representations through a learned per-scene codebook. 

We also include a comparison with the latest 3D unsupervised OCL based on NeRF, uOCF \cite{luo2024unsupervised} on the GSO dataset, as a baseline in the context of the network to contextualize the differences between implicit NeRF and explicit 3DGS representations for 3D OCL, providing qualitative and quantitative comparison.

\subparagraph{Metric}
We evaluate 3D scene OCL by comparing rendered feature masks derived from the learned Gaussian's identity features against ground-truth foreground object masks. We report clustering-based metrics, including Adjusted Rand Index (ARI) \cite{rand1971objective} and Adjusted Mutual Information (AMI) \cite{vinh2009information}. Both metrics are computed in two variants: foreground-only (FG-ARI / FG-AMI), which considers only object pixels, and all-pixel (ARI-A / AMI-A), which evaluates the entire scene. In addition, we report the mean Intersection over Union (mIoU) to assess the segmentation quality of the objects. All metrics are computed per scene under train view and reported as the mean across all test scenes. We also report reconstruction quality (PSNR, SSIM, LPIPS) and render speed (FPS) when comparing to uOCF.

\subparagraph{Implementation Details}
We train GOLD \cite{chen2025learning} separately on the training scenes of each dataset: 40 epochs for GSO and 30 epochs for OCTA, following\cite{chen2025learning}. We use $D_{slot}=128$ with 122 intrinsic and 6 extrinsic dimensions, $K=7$ maximum objects per scene, and $C=11$ total global objects. We implement the compared 3DGS baselines following their original pipelines, with SfM \cite{schonberger2016structure} point-cloud initialization. For Gaussian Grouping \cite{ye2024gaussian} and ObjectGS \cite{zhu2025objectgs},  we prepare the masks following their pipeline with SAM and DEVA, and prepared per-view SAM masks for Unified Lift \cite{zhu2025rethinking}. Our method is evaluated on both datasets with Gaussian identity features $f_i\in \mathbb{R}^{122}$. All methods are optimized for 300k iterations per-scene on train views.

We train uOCF \cite{luo2024unsupervised} following its original two-stage setup, initializing Stage 1 with the released pre-trained 3D object prior. For Stage 2, we train for 247k steps on the GSO dataset, then freeze the attention module and fine-tune only the decoder on each test scene for 1000 iterations to align with our method's OCL train split and per-scene optimization pipeline.

\begin{table*}[t]
\centering
\caption{Quantitative comparison to NeRF-based methods on the GSO dataset on OCL, reconstruction metrics, and render speed.}
\label{tab:NeRFmetric}
\centering
\resizebox{1\linewidth}{!}{
\begin{tabular}{ll|ccccccccc}
\toprule
Dataset & Method & FG-ARI $\uparrow$ & ARI-A $\uparrow$ & FG-AMI $\uparrow$ & AMI-A $\uparrow$ & mIoU $\uparrow$  &PSNR$\uparrow$  &SSIM$\uparrow$  &LPIPS $\downarrow$ &FPS$\uparrow$ \\
\midrule
\multirow{2}{*}{GSO}
& uOCF \cite{luo2024unsupervised} & 0.635 ± 0.281 &0.404 ± 0.097
 & 0.665 ± 0.264&0.356 ± 0.082& 0.508 ± 0.110 & 26.458 ± 3.337 & 0.835 ± 0.053 &\textbf{0.191 ± 0.073} & 6.2 ± 1.1 \\
 & Ours & \textbf{0.846 ± 0.208} & \textbf{0.644 ± 0.147} & \textbf{0.845 ± 0.204} & \textbf{0.558 ± 0.130} &\textbf{0.508 ± 0.130} & \textbf{30.007 ± 4.354}& \textbf{0.910 ± 0.041} & 0.246 ± 0.100 & \textbf{106.1 ± 54.3}  \\
\bottomrule
\end{tabular}
}
\end{table*}
\subsection{Experiment Results}
\subparagraph{OCTA Dataset} Fig.~\ref{fig:OverallVis}(a) shows a qualitative comparison on the OCTA dataset. Gaussian Grouping \cite{ye2024gaussian} and ObjectGS \cite{zhu2025objectgs} depend on SAM masks pre-processed with DEVA, so their learned Gaussian features remain strongly appearance-driven. Consequently, their rendered masks capture appearance similarity but often fail to preserve object-level consistency and internal part relationships. Unified Lift \cite{zhu2025rethinking} also uses SAM masks, but adopts a more complex end-to-end training scheme. Under limited-view supervision, this complexity can lead to instability and noisy Gaussian identity features.

In contrast, our pipeline uses the object-centric perception of GOLD to learn Gaussian identity features that encode both local appearance and coherent object structure. This yields more consistent segmentation of complete objects and better preservation of part relationships, producing more accurate and semantically meaningful decomposition (Table~\ref{tab:mainmetric}). Our method achieves the second-best mIoU while outperforming prior methods on OCL metrics. This mIoU–OCL tradeoff is due to GOLD’s DINO-based masks, which favor object-level coherence over the sharper boundaries typically produced by SAM.
 
\subparagraph{GSO Dataset}
In contrast to the OCTA dataset, the GSO dataset is synthetic; however, the increased complexity in object appearance leads to a higher likelihood of over-segmentation, coupled with less geometric information in the scene, resulting in overall lower metric scores for the VFM-based methods, as in Fig.~\ref{fig:OverallVis} (b) and Table~\ref{tab:mainmetric}. Furthermore, compared to OCTA, objects in GSO often exhibit different appearances across views, resulting in different identity features across views, making it more difficult for GOLD to learn discriminative object-centric representations, resulting in inconsistent identification across views (highlighted in the yellow box in Fig.~\ref{fig:discovery}).

Despite these challenges, our pipeline continues to outperform previous methods in OCL metrics and shows competitive segmentation results, as shown in Table~\ref{tab:mainmetric}. As a result, although our method still preserves coherent object-level structure, the overall metric is lower compared to the OCTA dataset.

\begin{figure}[t]
  \centering
  \includegraphics[width=1\linewidth]{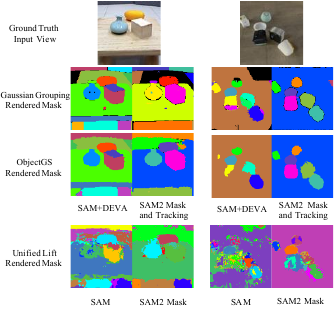}
  \caption{Qualitative comparison of rendered Gaussian feature masks using each prior method’s original supervision masks versus SAM2-based segmented/tracked mask supervision.}
  \label{fig:sam2}
\end{figure}
\begin{figure}[t]
  \centering
  \includegraphics[width=1\linewidth]{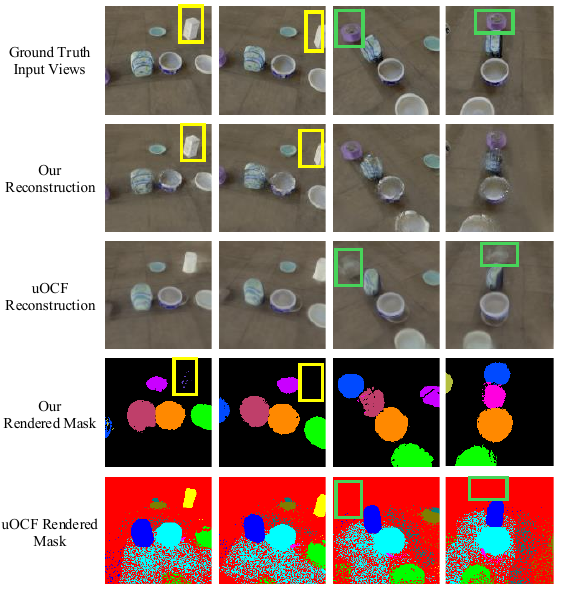}
  \caption{Qualitative comparison of reconstruction and rendered masks for our method and uOCF on the GSO dataset. As highlighted, when uOCF fails to discover an object, the reconstructed scene may miss that object, whereas our reconstruction remains intact even when object discovery is imperfect.}
  \label{fig:uOCF}
\end{figure}
\subparagraph{Cross Scene Identification}
With the learned global object codebook, our pipeline features cross-scene identification without the need for further re-training or fine-tuning of either the Gaussian identity features or codebook features. As shown in Fig.~\ref{fig:cross_scene}, as highlighted, Gaussian identity features learned under our global object-level codebook, the same object in different scenes holds the same ID, while other previous methods, in which the learned identification cannot be generalized across scenes, result in different IDs for the same object across different scenes.

\begin{table*}[t]
\centering
\caption{Quantitative comparison of our proposed pipeline and prior methods on the OCTA and GSO datasets, where the prior methods’ original supervision masks are replaced with SAM2-generated masks.}
\label{tab:mainmetricsam2}
\resizebox{0.9\textwidth}{!}{
\begin{tabular}{llccccc}
\toprule
Dataset & Method & FG-ARI $\uparrow$ & ARI-A $\uparrow$ & FG-AMI $\uparrow$ & AMI-A $\uparrow$ & mIoU $\uparrow$ \\
\midrule

\multirow{4}{*}{OCTA}
& Gaussian Grouping \cite{ye2024gaussian} + SAM2 mask and tracking
& \underline{0.839 ± 0.166} & \underline{0.322 ± 0.162} & \underline{0.826 ± 0.152} & \underline{0.520 ± 0.117} & \textbf{0.805 ± 0.099} \\
& Unified Lift \cite{zhu2025rethinking} + SAM2 mask
& 0.563 ± 0.116 & 0.249 ± 0.071 & 0.602 ± 0.116 & 0.390 ± 0.073 & 0.530 ± 0.088 \\
& ObjectGS \cite{zhu2025objectgs} + SAM2 mask and tracking
& 0.777 ± 0.177 & 0.209 ± 0.090 & 0.775 ± 0.159 & 0.385 ± 0.090 & 0.489 ± 0.154 \\
& Ours & \textbf{0.870 ± 0.061} & \textbf{0.760 ± 0.032} & \textbf{0.868 ± 0.042} & \textbf{0.677 ± 0.030} & \underline{0.593 ± 0.079} \\
\midrule

\multirow{4}{*}{GSO}

& Gaussian Grouping \cite{ye2024gaussian} + SAM2 mask and tracking
& \underline{0.826 ± 0.211} & \textbf{0.786 ± 0.177} & \underline{0.831 ± 0.207} & \textbf{0.744 ± 0.126} & \textbf{0.790 ± 0.083} \\
& Unified Lift \cite{zhu2025rethinking} + SAM2 mask
& 0.519 ± 0.180 & 0.605 ± 0.102 & 0.596 ± 0.190 & 0.500 ± 0.067 & \underline{0.532 ± 0.080} \\
& ObjectGS \cite{zhu2025objectgs} + SAM2 mask and tracking
& 0.667 ± 0.272 & 0.308 ± 0.281 & 0.730 ± 0.228 & 0.333 ± 0.196 & 0.306 ± 0.164 \\
& Ours & \textbf{0.846 ± 0.208} & \underline{0.644 ± 0.147} & \textbf{0.845 ± 0.204} & \underline{0.558 ± 0.130} & 0.508 ± 0.130 \\
\bottomrule
\end{tabular}
}
\end{table*}

\subparagraph{Comparing to NeRF-Based Methods}
We run both our method and uOCF \cite{luo2024unsupervised} on a single H200 GPU for all test scenes, and we show the comparison of the qualitative and quantitative results between our method and uOCF on the GSO dataset in Fig.~\ref{fig:uOCF} and Table~\ref{tab:NeRFmetric}, where our method outperforms uOCF in all metrics except LPIPS due to the smoothness of NeRF and perceptual loss in the training. This further emphasizes the advantage of 3DGS representation over NeRF, as 3DGS provides a more explicit representation, resulting in a higher quality of reconstruction and faster rendering. 

Furthermore, in NeRF-based approaches such as uOCF, OCL is entangled with scene reconstruction. As a result, the model’s ability to discover objects directly affects the reconstruction quality. As highlighted in Fig.~\ref{fig:uOCF}, when some objects are not discovered in the scenes, the reconstruction produced by uOCF is compromised, resulting in missing objects. In contrast, although our method may fail to discover certain objects, the overall scene reconstruction remains unaffected.

\subsection{Ablation Study}
To further isolate the effect of the supervision signal on object-centricity, we evaluate alternative supervision from prior VFM supervised methods by replacing the original SAM \cite{kirillov2023segment} and DEVA \cite{cheng2023tracking} processed masks with SAM2-generated masks \cite{ravi2024sam}, a more advanced VFM that also features mask tracking. Additionally, to promote object-centricity, we apply stricter post-processing by increasing the IoU confidence threshold to 0.8 and removing segments occupying less than 10\% of the image area.

\begin{figure}[t]
  \centering
  \includegraphics[width=0.95\linewidth]{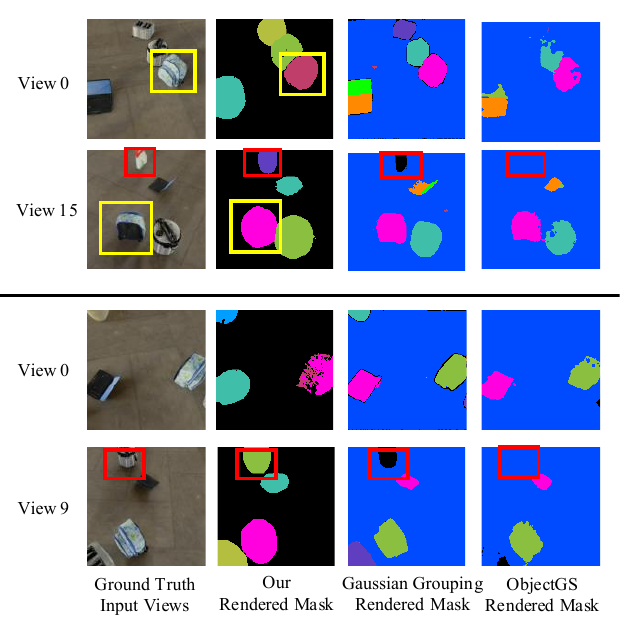}
  \caption{Object discovery capability of our method (red boxes). Although SAM2 masks and tracking improve object-centric supervision, they fail to generalize to newly appearing objects that are absent in the first view without manual re-prompting.}
  \label{fig:discovery}
\end{figure}

As shown in Fig.~\ref{fig:sam2} and Table~\ref{tab:mainmetricsam2}, SAM2 supervision substantially improves object-centricity both visually and quantitatively. Gaussian Grouping \cite{ye2024gaussian} outperforms our method on ARI-A and AMI-A, due to the sharper masks SAM2 produces that separate foreground objects more cleanly. However, despite this improvement, SAM2-based supervision still lacks cross-scene identity generalization and object discovery. As shown in Fig.~\ref{fig:discovery}, in GSO scenes where some objects are absent from the first input view, SAM2’s memory-based tracking fails to identify newly appearing objects without manual re-prompting. In contrast, our pipeline generalizes across scenes and consistently discovers and identifies newly appearing objects for supervision without manual intervention, highlighting the advantage of object-centric supervision over VFM-derived signals.

\section{Limitation}
Our method is currently limited by the scarcity of datasets that support both OCL and 3DGS, restricting scalability to complex real-world scenes with dense interactions and high variability. Training GOLD also poses practical challenges, including choosing the number of slots and handling the higher computational cost of larger codebooks, which makes training more resource-intensive. We view these limitations as stemming primarily from current data and computing constraints rather than the method itself. With larger, more diverse real-world datasets and sufficient computational resources, GOLD could learn a more expressive global object codebook that generalizes across scenes and supports a universal object-centric representation for supervising 3DGS.
\section{Conclusion}
We introduced a novel supervising method for learning identity features in 3DGS. Compared to prior VFM-based method, our proposed method learns global object-centric identity features without the need for any pre-/post-process of the mask and pipeline, nor a tailored training process and loss design. The rendered feature mask of the learned global identity feature outperforms the prior method in OCL metrics with competitive segmentation quality under limited views. We further demonstrate that our method achieves cross-scene identification and object discovery capabilities that are lacking when recent advanced VFMs are used as supervision.
{
    \small
    \bibliographystyle{ieeenat_fullname}
    \bibliography{main}
}


\end{document}